\pdfoutput=1

\documentclass[11pt]{article}

\usepackage[]{acl}

\usepackage{times}
\usepackage{latexsym}

\usepackage[T1]{fontenc}

\usepackage[utf8]{inputenc}

\usepackage{microtype}

\usepackage{inconsolata}
\usepackage{graphicx}
\usepackage{xcolor}
\usepackage{amsmath}
\usepackage{makecell}
\usepackage{subcaption}

\title{Polarity Calibration for Opinion Summarization}

\author{Yuanyuan Lei\textsuperscript{1}, Kaiqiang Song\textsuperscript{2}, Sangwoo Cho\textsuperscript{2}, Xiaoyang Wang\textsuperscript{2}, \\
        \vspace{3pt}
        \bf{Ruihong Huang\textsuperscript{1}, Dong Yu\textsuperscript{2}} \\
        \vspace{1pt}
        \textsuperscript{1}Texas A\&M University \quad \textsuperscript{2}Tencent AI Lab, Bellevue, WA \\
        \vspace{1pt}
        \texttt{\{yuanyuan, huangrh\}@tamu.edu}\\
        \texttt{\{riversong, swcho, shawnxywang, dyu\}@global.tencent.com}}

\begin{document}
\maketitle
\begin{abstract}

Opinion summarization is automatically generating summaries from a variety of subjective information, such as product reviews or political opinions. The challenge of opinions summarization lies in presenting divergent or even conflicting opinions. We conduct an analysis of previous summarization models, which reveals their inclination to amplify the polarity bias, emphasizing the majority opinions while ignoring the minority opinions. To address this issue and make the summarizer express both sides of opinions, we introduce the concept of polarity calibration, which aims to align the polarity of output summary with that of input text. Specifically, we develop a reinforcement training approach for polarity calibration. This approach feeds the polarity distance between output summary and input text as reward into the summarizer, and also balance polarity calibration with content preservation and language naturality. We evaluate our \underline{Po}larity \underline{Ca}libration model (\textit{PoCa}) on two types of opinions summarization tasks: summarizing product reviews and political opinions articles. Automatic and human evaluation demonstrate that our approach can mitigate the polarity mismatch between output summary and input text, as well as maintain the content semantic and language quality\footnote{The code and data link: https://github.com/yuanyuanlei-nlp/polarity\_calibration\_naacl\_2024}.

\end{abstract}

\section{Introduction}

Opinions are prevalent in various areas, such as social media posts, customer reviews, spoken conversations, argumentative debates, or political matters \cite{pang2008opinion, liu2022sentiment}. Opinion summarization enables automatically generating a brief and informative summary from a large volume of opinions, reviews, or subjective text \cite{hu2004mining, ganesan2010opinosis, lei-huang-2022-shot, 10008998, angelidis2018summarizing, amplayo-lapata-2021-informative}. The automatic opinion summarization models simplify the extraction of valuable insights from the extensive pool of subjective content, playing a pivotal role in various information access applications, such as digest creation, decision making, product development, or public perception monitoring \cite{suhara-etal-2020-opiniondigest, amplayo-etal-2021-aspect, iso-etal-2022-comparative}.

The challenge of opinion summarization lies in presenting divergent or even conflicting opinions. This contrasts sharply with summarizing objective content such as government reports, scientific research, or legal documents, which typically present factual information without the layer of personal perspectives \cite{erera-etal-2019-summarization, kornilova-eidelman-2019-billsum, cachola-etal-2020-tldr, cao-wang-2022-hibrids}. Take summarizing product reviews as an example, customers often express differing opinions about the same product, including both positive and negative viewpoints. The central challenge of subjective summarization is aggregating and presenting these disparate opinions.

\begin{figure*}[ht]
  \centering
  \includegraphics[width = 4.3in]{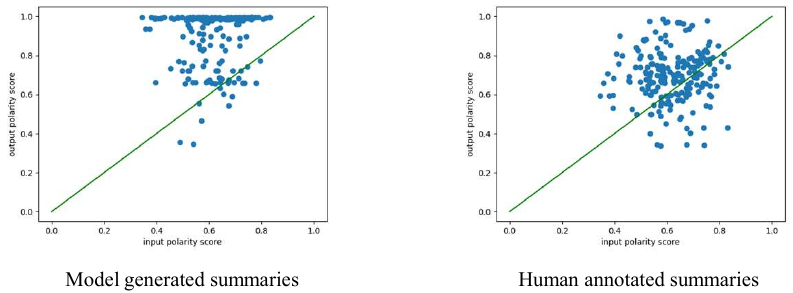}
  \caption{The x-axis represents input text polarity score, and the y-axis represents output summary polarity score. The model can amplify the polarity bias, by presenting the majority opinions while ignoring the minority opinions.}
  \label{polarity_amplify}
\end{figure*}

The critical observation of previously developed summarization models is their tendency to amplify the polarity bias of input text, presenting the majority opinions while ignoring the minority opinions (Section \ref{section_analysis}). In the example of summarizing product reviews, we quantify the polarity scores of input text and output summaries. Our findings reveal that when the majority of customers express positive opinions about a product, the summarization models directly trained on overwhelming positive reviews can be easily biased to generate overly positive summaries while neglecting the minority of negative opinions (Figure \ref{polarity_amplify}, \ref{visualization_analysis}). The amplification of polarity bias indicates the limitation in the previous approaches.

To address this issue and proportionally express both sides of opinions, we propose the idea of \textit{polarity calibration}, which aims to align the polarity of output summary with that of input text. In contrast to previous work, we argue that when dealing with conflicting opinions, an intelligent summarizer should proportionally present both sides of majority and minority opinions, and align with the polarity of input text. Thus, we propose to integrate an additional layer of polarity calibration guidance into the summarizer. The objective of polarity calibration is to encourage the summarizer to exhibit both sides of viewpoints, and mitigate the polarity mismatch between output and input.

To achieve polarity calibration, we develop a reinforcement training approach. More specifically, we employ a polarity reward model to assess the polarity distance between output summary and input text. The polarity distance is incorporated into the summarization model as a reward signal, to encourage the minimization of polarity discrepancy. Besides, to guide the summarizer to maintain the original semantic content of input text, we train a content preservation reward model and feed the content similarity between output and input as reward into the summarizer. In addition, to promote the generation of naturally flowing language, we employ a language naturality reward model and leverage language fluency score as reward. By aggregating the rewards for polarity distance, content preservation, and language naturality, the reinforcement training is designed to balance between improving polarity alignment, retaining content semantic, and generating fluent language.

We evaluate our approach on two types of opinions summarization tasks: summarizing product reviews and political opinions articles. The experiments on both two tasks demonstrate the effectiveness of our method in decreasing the polarity discrepancy between output and input. Both automatic and human evaluation confirm that our approach can enhance polarity alignment, while maintaining content semantic and language quality. Our main contributions are summarized as follows:

\begin{itemize}
    \item Motivated by the analysis that opinion summarizers tend to amplify the polarity bias, we firstly propose \textit{polarity calibration}, to align the polarity of output summary and input text.
    \item We design a reinforcement training approach to achieve polarity calibration, by integrating the three rewards for polarity distance, content preservation, language naturality.
    \item We conduct experiments on two opinions summarization tasks, and effectively decrease the polarity distance while maintaining content semantic and language fluency.
\end{itemize}

\section{Polarity Bias Amplification}

\label{section_analysis}

This section provides a quantitative analysis of previous summarization models, which reveals their tendency to amplify the polarity bias.

Take product reviews summarization as an example, we aim to examine the polarity of output summary and input reviews. To quantify the polarity, we train a sentiment analysis model on the Amazon product reviews dataset \cite{marc_reviews} to generate polarity scores. This sentiment analyzer is trained to predict whether a review sentence is positive or negative, and we use the predicted probability of the positive class as the polarity score. The polarity score is a numerical value on a scale from zero to one, with zero indicating extreme negative and one indicating extreme positive. Figure \ref{polarity_amplify} illustrate the polarity analysis on the Amazon product reviews summarization dataset (AmaSum) \cite{brazinskas-etal-2021-learning, hosking-etal-2023-attributable}. The x-axis is the average polarity score of sentences in customer reviews, and y-axis is the average polarity score of sentences in summary. Each blue data point represents one product.

\begin{figure*}[ht]
  \centering
  \includegraphics[width = 6.3in]{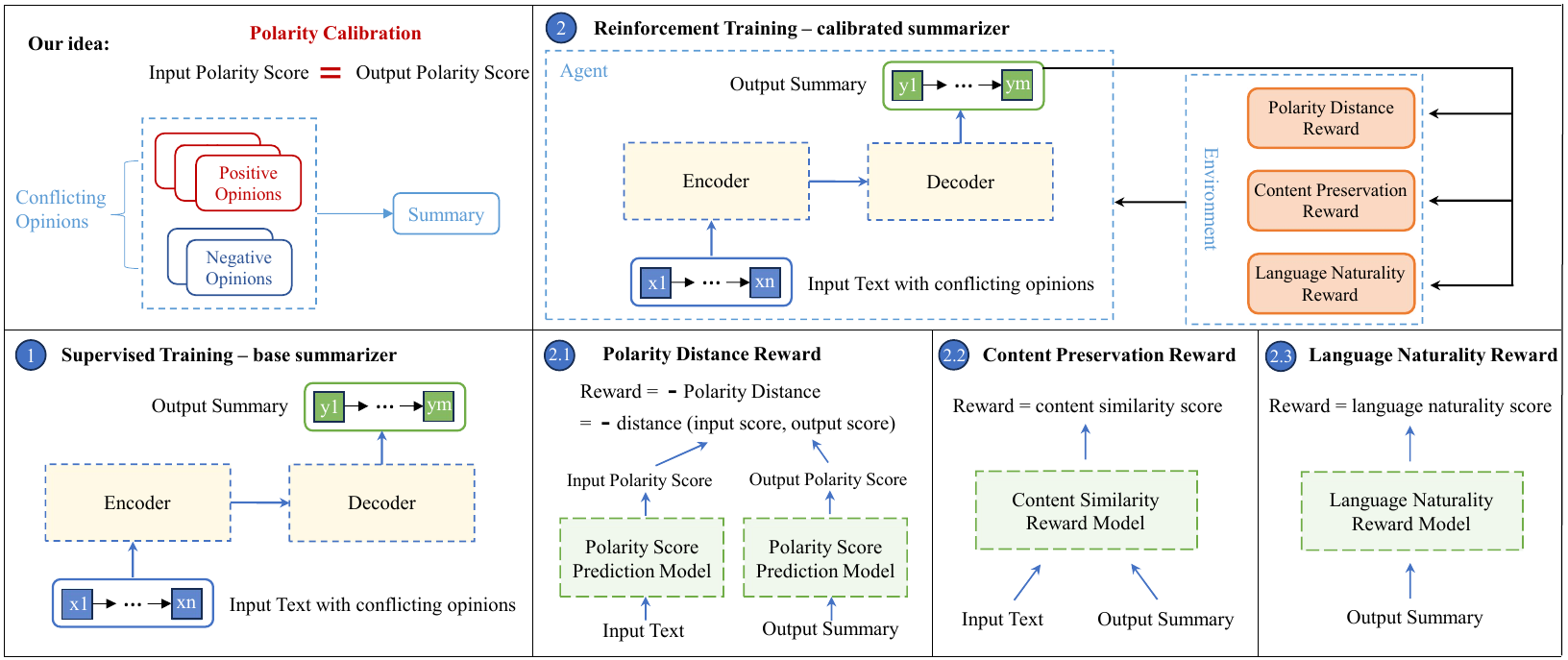}
  \caption{An illustration of polarity calibration with reinforcement learning.}
  \label{method_figure}
\end{figure*}

The comparison between model generated summaries and human annotated summaries reveals the model's inclination to magnify polarity bias. Figure \ref{polarity_amplify} takes CopyCat model \cite{brazinskas-etal-2020-unsupervised} as an example, but this observation also exists in other summarization models (Figure \ref{visualization_analysis}). While human consciously maintain the polarity level of input text when crafting summaries, the models tend to generate overly positive summaries, overlooking the minority of negative opinions. One possible explanation is that the models trained on the text predominated with positive reviews tend to develop a bias in favor of highlighting the majority of positive opinions. To guide the model to present both sides of opinions in a proportional manner and better align with the input, we propose to calibrate the polarity score of output summary and input text.

\section{Polarity Calibration}

This section introduces the methodology for polarity calibration, which is illustrated in Figure \ref{method_figure}. The polarity calibration is designed in two steps: firstly training a base summarizer with supervised learning, to equip the model with opinion summarization ability, and secondly training a calibrated summarizer through reinforcement learning, with the aim of refining the model's polarity alignment. The calibrated summarizer after \underline{po}larity \underline{ca}libration is named as \textit{PoCa}.

\subsection{Supervised Training}

In the supervised training stage, we train a base summarizer $M_{base}$, with the ability to summarize opinions. We employ the flan-T5-large model as the backbone model \cite{chung2022scaling, raffel2023exploring}. The input $x$ is the concatenation of different reviews or opinions, denoted as $(x_1, x_2, ..., x_n)$. The goal of the base summarizer is to generate a summary $y=(y_1, y_2, ..., y_m)$, and the human annotated summaries serve as the ground truth labels for training. The cross-entropy loss (CE) is utilized as the learning objective:
\begin{equation}
    L_{CE} = \sum_{t=1}^T \log \pi_\theta(y_t | y_{t-1}, x)
\end{equation}

The supervised training makes the base summarizer generate text that is close to the human written reference. However, simply minimizing cross-entropy loss without additional polarity knowledge does not guarantee the polarity alignment between output and input. A generated text that only express the majority opinion can also achieve high Rouge score when compared to human written reference. To imbue the model with polarity awareness, we propose to incorporate an extra guidance of polarity calibration through reinforcement learning.

\subsection{Reinforcement Training}

In the reinforcement training stage, we train a calibrated summarizer $M_{calibrate}$ on the basis of base summarizer $M_{base}$. The input $x$ is the concatenation of different reviews or opinions. The goal of the calibrated summarizer is to generate a summary $\hat{y}=(\hat{y}_1, \hat{y}_2, ..., \hat{y}_m)$ that retains the semantics of the input $x$ and calibrates with its polarity level.

We formulate the reinforcement learning for polarity calibration as a system composed of an agent (A), action ($a$), policy ($\pi$), and reward ($R$). The agent is the summarization model with parameters $\theta$ that observes the current state (the model output) at time $t$ and takes an action $a$ (predict the next word $\hat{y}_t$) by using a policy ($\pi$). The reward ($R$) is a scalar calculated by the reward models $R: \hat{y} \rightarrow [0,1]$, to evaluate the quality of generated text $\hat{y}$. This reward is then returned as feedback to the summarization model. The objective of reinforcement learning is to maximize the reward ($R$) by updating the parameters $\theta$ of the agent:
\begin{equation}
    J(\theta) = E_{\pi_\theta(\hat{y}|x)}[R(\hat{y})]
\end{equation}

Since the reward is the discrete function of the model's output, the reinforcement learning objective $J(\theta)$ is non-differentiable with respect to the model parameter $\theta$, which makes it difficult to back-propagate the error signals from the reward models to the summarizer. This issue can be addressed through policy gradient \cite{sutton1999policy}. Specifically, the expected reward is approximated using a sampling method and the model is trained using stochastic gradient ascent \cite{williams1992simple}, which can be formulated as:
\begin{equation}
    \bigtriangledown_\theta J(\theta) = E_{\pi_\theta(\hat{y}|x)}[R(\hat{y}) \bigtriangledown_\theta \log_{\pi_\theta}(\hat{y}|x)]
\end{equation}
where $\pi_\theta$ is a policy that generates a probability of picking a word as output. The policy gradient learns the optimal policy directly by modifying the model parameters based on the observed rewards.

\subsection{Reward Models}

In the reinforcement training stage, the generated summary $\hat{y}$ is expected to meet three objectives: (i) reduce polarity distance between output and input (ii) preserve the content semantics of input text (iii) ensure language to be grammatical correct and fluent. Based on the above objectives, the designed reward function $R(x, \hat{y})$ consists of three rewards:
\begin{equation}
    R(x, \hat{y}) = \alpha R_P(x, \hat{y}) + \beta R_C(x, \hat{y}) + \gamma R_L(\hat{y})
\end{equation}
where $R_P(x, \hat{y})$ is the polarity distance reward calculated between output $\hat{y}$ and input $x$, $R_C(x, \hat{y})$ is the content semantic similarity reward between output $\hat{y}$ and input $x$, and $R_L(\hat{y})$ is the language fluency reward for the output text $\hat{y}$. The hyper-parameter $\alpha, \beta, \gamma$ represent weights for the respective rewards.

\subsubsection{Polarity Distance Reward}

The purpose of polarity distance reward is to minimize the polarity difference between output summary $\hat{y}$ and input text $x$. To measure the polarity distance, we build a polarity score prediction model that quantifies the polarity level of a given text. The polarity distance reward $R_P(x, \hat{y})$ is defined as the negative difference between polarity scores of output summary $\hat{y}$ and input text $x$:
\begin{equation}
    R_P(x, \hat{y}) = - \lvert polarity (\hat{y}) - polarity (x) \rvert
\end{equation}

The polarity score prediction model is tailored to accommodate different tasks. This paper explores two types of opinion summarization tasks: summarizing product reviews with positive or negative opinions, and summarizing political articles with liberal or conservative political stances.

For the Amazon product reviews summarization task, the polarity score prediction model is a sentiment analysis model. A binary classifier is built based on RoBERTa \cite{liu2019roberta}, to categorize a review sentence into positive or negative. The classifier is trained on the Amazon product reviews dataset \cite{marc_reviews}. The polarity score is the predicted probability of the positive class. Since the input text consists of multiple review sentences, the polarity score of input text $x$ is computed as the average of polarity scores assigned to individual review sentence. The polarity score of output summary $\hat{y}$ is computed as the average polarity score of sentences in the summary.

For the political articles summarization task, the polarity score prediction model is a political stance prediction model. A binary classifier is built based on RoBERTa \cite{liu2019roberta}, to categorize each article into liberal or conservative stance. The classifier is trained on the political stance dataset AllSides \cite{baly-etal-2020-detect}. The polarity score of each article is the predicted probability of the conservative class. Given the input text comprising multiple articles with different stances, the polarity score of input text $x$ is computed as the average of polarity scores assigned to each individual article. The polarity score of output summary $\hat{y}$ is also the predicted probability of conservative class for the summary.

\subsubsection{Content Preservation Reward}

The content preservation reward aims to ensure that the information expressed in the input text is retained in the output summary. To quantify the level of content preservation, we build a content similarity reward model to predict the similarity score between output $\hat{y}$ and input $x$. A RoBERTa based model is used that takes the $(x, \hat{y})$ pair as input and produce a similarity score. This content similarity reward model is trained on the STS-B semantic similarity dataset \cite{wang2018glue}. Considering the raw predicted similarity score ranges from zero to five, we normalize this raw score into the scale of zero to one, and define it as the reward.
\begin{equation}
   R_C(x, \hat{y}) = similarity (\hat{y}, x)
\end{equation}

\subsubsection{Language Naturality Reward}

The language naturality reward encourages the generated summary $\hat{y}$ to be grammatically correct and natural sounding. To assess the language naturality, we build a language fluency reward model that predicts the language fluency score of the output $\hat{y}$. A binary classifier using RoBERTa is built to predict the generated summary $\hat{y}$ into grammatical correctness or not. This language naturality reward model is trained on the Corpus of Linguistic Acceptability (CoLA) dataset \cite{warstadt2018neural}. The language naturality reward is defined as the predicted probability of the grammatical correctness class.
\begin{equation}
   R_L(\hat{y}) = fluency (\hat{y})
\end{equation}

\section{Experiments}

\subsection{Datasets}

We evaluate our approach using two datasets, each focusing on different types of opinions.

\noindent\textbf{AmaSum} \cite{brazinskas-etal-2021-learning} is the Amazon product reviews summarization dataset, which includes product reviews from a wide range of categories. We use the version that contains a maximum of 100 reviews per product for experiments. The dataset collects human written summaries from professional review websites. The annotated summary consists of three portions: verdicts that emphasize the most important points about a product, pros that describe positive details, and cons that states negative aspects. These three portions are concatenated together to form a single summary. We follow the previous work \cite{hosking-etal-2023-attributable} to evaluate the model on the testing set, which contains 50 products from each of the following four common categories: Electronics, Home \& Kitchen, Shoes, Sports \& Outdoors.

\vspace{3pt}

\noindent\textbf{NeuS} \cite{lee-etal-2022-neus} is the political opinions articles summarization dataset, which collects US political news articles from AllSides website. The articles with different political stances that discuss the same event are grouped together as a cluster. Each cluster contains three articles. The dataset also provides an expert written summary for each cluster of articles. We follow the dataset splitting setting released by \citet{lee-etal-2022-neus}, which results in 2452 / 307 / 307 news clusters allocated to the train, development, and test sets, respectively.

\begin{table*}[t]
    \centering
    \scalebox{0.87}{
    \begin{tabular}{|l||cc||cccc|}
        \hline
         & \multicolumn{2}{c||}{Polarity Distance} & \multicolumn{4}{c|}{Rouge Scores} \\
         & RMSE & MAE & Rouge-1 & Rouge-2 & Rouge-L & Rouge-Lsum \\
        \hline
        Human annotated summaries & 0.1794 & 0.1409 & - & - & - & - \\
        LexRank \cite{erkan2004lexrank} & 0.2772 & 0.2442 & 19.91 & 2.61 & 12.09 & 18.27 \\
        CopyCat \cite{brazinskas-etal-2020-unsupervised} & 0.3264 & 0.2907 & 17.38 & 1.36 & 10.95 & 15.80 \\
        BiMeanVAE-avg \cite{iso-etal-2021-convex-aggregation} & 0.2819 & 0.2549 & 21.31 & 2.00 & 12.32 & 19.63 \\
        BiMeanVAE-COOP \cite{iso-etal-2021-convex-aggregation} & 0.2537 & 0.2189 & 23.67 & 2.71 & 13.96 & 21.66 \\
        QT \cite{angelidis-etal-2021-extractive} & 0.2091 & 0.1609 & 21.17 & 1.55 & 11.36 & 19.53 \\
        SemAE \cite{basu-roy-chowdhury-etal-2022-unsupervised} & 0.2285 & 0.1786 & 20.32 & 1.62 & 11.35 & 18.60 \\
        Hercules-abstractive \cite{hosking-etal-2023-attributable} & 0.2469 & 0.2167 & 19.82 & 2.15 & 11.71 & 18.95 \\
        Hercules-extractive \cite{hosking-etal-2023-attributable} & 0.1888 & 0.1556 & 22.89 & 3.07 & 12.55 & 21.44 \\
        ChatGPT (gpt-3.5-turbo) & 0.2272 & 0.1875 & 23.31 & 2.76 & 12.99 & 21.32 \\
        GPT-4 (gpt-4) & 0.2005 & 0.1749 & 23.06 & 2.60 & 12.31 & 21.08 \\
        \hline
        base summarizer (flan-T5-large) & 0.2154 & 0.1782 & \textbf{29.23} & \textbf{5.64} & \textbf{17.19} & \textbf{26.69} \\
        calibrated summarizer (PoCa) & \textbf{0.1824} & \textbf{0.1533} & 28.44 & 5.12 & 16.96 & 25.92 \\
        \hline
    \end{tabular}}
    \caption{Automatic Evaluation of product reviews summarization on AmaSum dataset. The root mean squared error and mean absolute error between input text polarity score and output summary polarity score are reported.}
    \label{amasum_result}
\end{table*}

\begin{table*}[t]
    \centering
    \scalebox{0.87}{
    \begin{tabular}{|l||cc||cccc|}
        \hline
         & \multicolumn{2}{c||}{Polarity Distance} & \multicolumn{4}{c|}{Rouge Scores} \\
         & RMSE & MAE & Rouge-1 & Rouge-2 & Rouge-L & Rouge-Lsum \\
        \hline
        Human annotated summaries & 0.1984 & 0.1517 & - & - & - & - \\
        LexRank \cite{erkan2004lexrank} & 0.2282 & 0.1838 & 38.68 & 15.94 & 25.66 & 33.67  \\
        BART \cite{lewis-etal-2020-bart} & 0.2799 & 0.2291 & 38.22 & 15.73 & 25.52 & 34.24 \\
        Pegasus \cite{zhang2020pegasus} & 0.2810 & 0.2344 & 37.33 & 16.02 & 25.54 & 31.45 \\
        NeuS \cite{lee-etal-2022-neus} & 0.2172 & 0.1666 & 39.09 & 18.93 & 29.73 & 35.35 \\
        ChatGPT (gpt-3.5-turbo) & 0.2552 & 0.2076 & 42.01 & 16.24 & 26.12 & 37.27 \\
        GPT-4 (gpt-4) & 0.2626 & 0.2133 & 42.35 & 16.48 & 26.30 & 37.30 \\
        \hline
        base summarizer (flan-T5-large) & 0.2162 & 0.1613 & \textbf{43.83} & \textbf{20.75} & 31.75 & \textbf{39.16}\\
        calibrated summarizer (PoCa) & \textbf{0.1834} & \textbf{0.1389} & 43.68 & 20.70 & \textbf{31.98} & \textbf{39.16} \\
        \hline
    \end{tabular}}
    \caption{Automatic Evaluation of political opinions articles summarization on NeuS dataset. The root mean squared error and mean absolute error between input text polarity score and output summary polarity score are reported.}
    \label{neus_result}
\end{table*}

\subsection{Baselines}

Summarizing product reviews has attracted research attention for years. The following models are previously developed for summarizing product reviews and implemented as our baselines:

\noindent\textbf{CopyCat} \cite{brazinskas-etal-2020-unsupervised} is an abstractive method by using the hierarchical continuous latent representations to model products and reviews.

\noindent\textbf{BiMeanVAE} \cite{iso-etal-2021-convex-aggregation} is an abstractive method that encode full reviews as continuous latent vectors, by taking the average or optimizing the combination of review embeddings (COOP).

\noindent\textbf{QT} \cite{angelidis-etal-2021-extractive} uses vector quantization to map sentences to a discrete encoding space, and generates extractive summaries by selecting representative sentences from clusters.

\noindent\textbf{SemAE} \cite{basu-roy-chowdhury-etal-2022-unsupervised} is an extractive method that extends the QT method, by relaxing the discretization and encoding sentences as mixtures of learned embeddings.

\noindent\textbf{Hercules} \cite{hosking-etal-2023-attributable} develops both extractive and abstractive method, by encoding sentences from customer reviews into a hierarchical latent space and identifying common opinions.

\vspace{6pt}

\noindent Summarizing political articles with diverse political opinions has a relatively short research history. There are few previously established methods available for comparison. We follow \citet{lee-etal-2022-neus} to compare with the following systems:

\noindent\textbf{LexRank} \cite{erkan2004lexrank} is an unsupervised extractive graph-based model that selects
sentences based on graph centrality. The nodes are sentences and the edges are weighted with tf-idf.

\noindent\textbf{BART} \cite{lewis-etal-2020-bart} is a multi-document summarization model that fine tunes BART-large on the Multi-News dataset \cite{fabbri-etal-2019-multi}.

\noindent\textbf{Pegasus} \cite{zhang2020pegasus} is an abstractive model that fine tunes Pegasus-large model on the Multi-News dataset \cite{fabbri-etal-2019-multi}.

\noindent\textbf{NeuS} \cite{lee-etal-2022-neus} develops an abstractive summarization method that learns to generate summary in a hierarchical order from title to article.

\noindent\textbf{ChatGPT} is a large language model that generates abstractive summaries via prompting. We use the gpt-3.5-turbo version to obtain the summary.

\noindent\textbf{GPT-4} is another large language model that automatically generates abstractive summaries. We use the gpt-4 version to create the summaries. The prompt provided to the model is in Appendix \ref{gpt-prompt}.

\subsection{Automatic Evaluation}

The automatic evaluation metrics for polarity calibration are the root mean squared error (RMSE) and mean absolute error (MAE) between polarity scores of output summary and input text. The evaluation metrics for content semantics are calculated by the Rouge scores \cite{lin-2004-rouge} between model generated summaries and human written reference. The expectation for the summarizer is to minimize polarity distance while maximizing Rouge scores. The results for product reviews summarization on AmaSum dataset are presented in Table \ref{amasum_result}. The results for political opinions articles summarization on NeuS dataset are shown in Table \ref{neus_result}. Our \underline{po}larity \underline{ca}libration model is named as \textit{PoCa}, and is reported in the last row of tables.

The results demonstrate that polarity calibration through reinforcement training can effectively reduce polarity distance between generated summary and input text while preserving content semantics. When compared to the base summarizer, the calibrated summarizer (PoCa) consistently reduces polarity distance on both AmaSum and NeuS datasets. This indicates that our approach successfully improves polarity alignment between output and input, by incorporating polarity calibration as additional guidance. The statistical t-test indicates a significant difference between calibrated summarizer and base summarizer in terms of polarity distance, but no significant difference in terms of Rouge scores, under the confidence level of 95\%. This proves the effectiveness of our approach in mitigating polarity bias without compromising on content semantics.

\begin{table*}[t]
    \centering
    \scalebox{0.87}{
    \begin{tabular}{|l||cc||cccc|}
        \hline
         & \multicolumn{2}{c||}{Polarity Distance} & \multicolumn{4}{c|}{Rouge Scores} \\
         & RMSE & MAE & Rouge-1 & Rouge-2 & Rouge-L & Rouge-Lsum \\
        \hline
        base summarizer & 0.2154 & 0.1782 & \textbf{29.23} & \textbf{5.64} & \textbf{17.19} & \textbf{26.69} \\
        + polarity reward & \textbf{0.1545} & \textbf{0.1247} & 25.24 & 4.70 & 15.71 & 23.05 \\
        + polarity + content reward & 0.1839 & 0.1547 & 28.13 & 5.22 & 16.68 & 25.78 \\
        + polarity + content + language reward & 0.1824 & 0.1533 & 28.44 & 5.12 & 16.96 & 25.92 \\
        \hline
    \end{tabular}}
    \caption{The ablation study of three rewards in reinforcement training on AmaSum dataset.}
    \label{ablation_study}
\end{table*}

\begin{table*}[t]
    \centering
    \scalebox{0.84}{
    \begin{tabular}{|l||c||cc||cc|}
        \hline
        & \multicolumn{1}{c||}{Polarity} & \multicolumn{2}{c||}{Content} & \multicolumn{2}{c|}{Language} \\
        & Polarity Distance & Non-hallucination & Non-redundancy & Fluency & Coherency \\
        \hline
        QT \cite{angelidis-etal-2021-extractive} & 0.425 & 0.70 & \textbf{0.80} & 0.30 & 0.15 \\
        Hercules \cite{hosking-etal-2023-attributable} & 0.450 & 0.90 & 0.25 & 0.70 & 0.15 \\
        ChatGPT (gpt-3.5-turbo) & 0.425 & 0.90 & 0.65 & 0.90 & 0.90 \\
        \hline
        base summarizer (flan-T5-large) & 0.450 & 0.85 & 0.75 & \textbf{0.95} & 0.85 \\
        calibrated summarizer (PoCa) & \textbf{0.350} & \textbf{0.95} & \textbf{0.80} & \textbf{0.95} & \textbf{0.95} \\
        \hline
    \end{tabular}}
    \caption{Human Evaluation of polarity bias, content semantics, and language quality on AmaSum dataset.}
    \label{human_eval}
\end{table*}

\subsection{Ablation Study}

This section studies the effect of three rewards in reinforcement training, and the results on AmaSum dataset is shown in \ref{ablation_study}. We observe that only feeding the polarity distance reward into the summarizer can achieve the lowest polarity distance, however, the content semantics is compromised compared to the base summarizer. Learning from both the polarity distance reward and content preservation reward jointly can lead to a reduction in polarity distance while also retaining content semantics. Incorporating the three rewards together can strike a balance between polarity calibration and maintaining content semantics and language quality. This suggests that the three rewards are essential for refining summarization models.

\subsection{Human Evaluation}

The human evaluation aims to assess the generated summaries from three perspectives: polarity bias, content semantics, and language quality. Specifically, we provide the human annotators with input text and model generated summaries, and ask them five questions regarding polarity bias, content non-hallucination, content non-redundancy, language fluency, and language coherency (Appendix \ref{human_eval_questions}). Four human annotators who are specialized in natural language processing participated in the evaluation. We select three baselines that achieve low polarity distance, QT \cite{angelidis-etal-2021-extractive}, Hercules \cite{hosking-etal-2023-attributable}, and ChatGPT, along with our base summarizer and calibrated summarizer for evaluation. To avoid the leakage of model information, different models are randomly shuffled and the name of models are omitted. After collecting the evaluation scores from human annotators for the five questions , we normalize the metrics into the range of zero to one. For the polarity bias evaluation, a lower score is preferable, whereas for the other four questions related to content and language, a higher score is better. The results of human evaluation on AmaSum dataset are presented in Table \ref{human_eval}.

The human evaluation showcases the consistent observations with automatic evaluation. The calibrated summarizer exhibits the lowest polarity bias compared to the other models. This proves that the polarity calibration through reinforcement training is effective in mitigating polarity bias. Besides, the calibrated summarizer also achieve high scores for content semantics and language quality. This indicates that the design of content preservation reward and language naturality reward has the effect of improving content and language quality. In summary, our approach enables the mitigation of polarity bias, and also enhances content preservation and language naturality.

\subsection{Qualitative Analysis}

\begin{figure*}[ht]
  \centering
  \includegraphics[width = 6.3in]{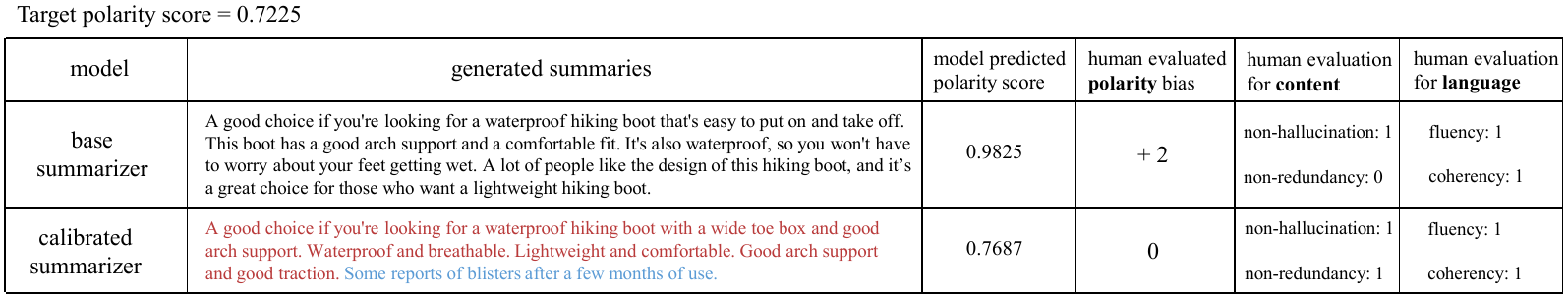}
  \caption{The qualitative analysis of generated summaries from base summarizer and calibrated summarizer.}
  \label{qualitative_analysis}
\end{figure*}

\begin{figure*}[ht]
  \centering
  \includegraphics[width = 6.3in]{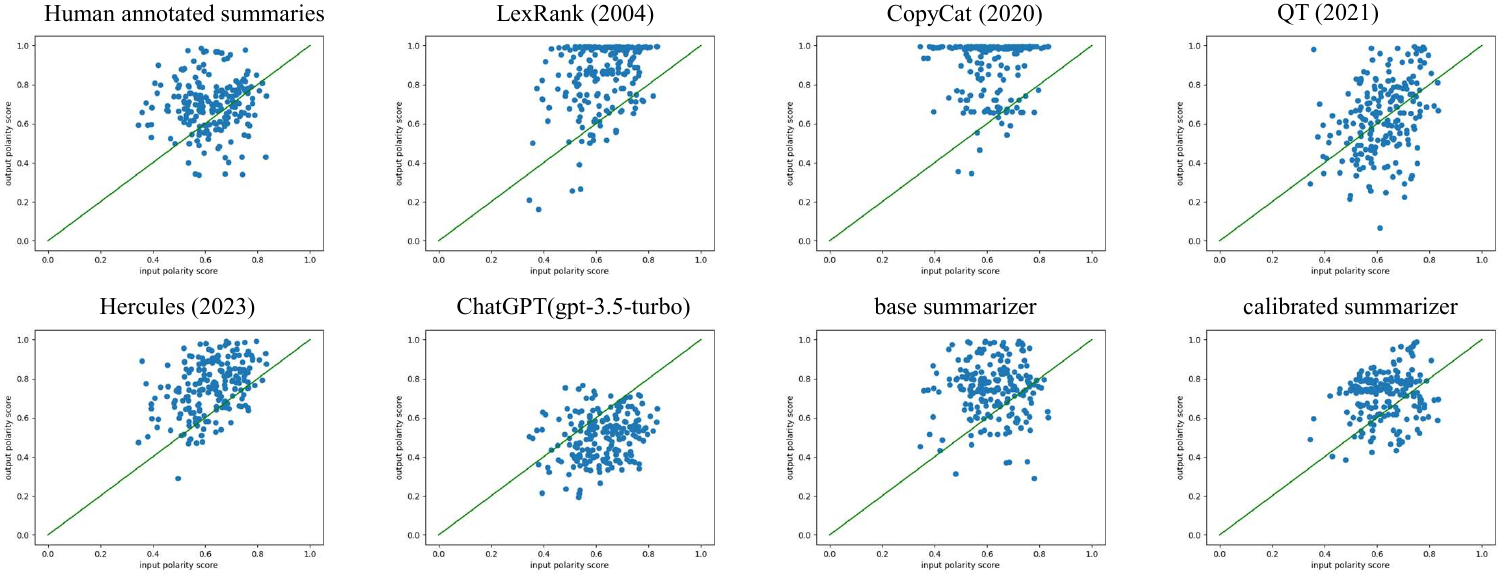}
  \caption{The visualization analysis of generated summaries from various models. The x-axis is input text polarity score, and the y-axis is output summary polarity score. The ideal polarity calibration lies on the green y=x line.}
  \label{visualization_analysis}
\end{figure*}

Figure \ref{qualitative_analysis} shows an example of generated summaries from base summarizer and calibrated summarizer. Figure \ref{qualitative_analysis_full} showcases the generated summaries from additional models. We observe that the summary generated by the base summarizer only includes positive opinions without mentioning any negative opinions. This corroborates our finding that the model without polarity calibration tends to develop a bias of emphasizing the majority opinion while overlooking the minority opinion. Nevertheless, the summary produced by the calibrated summarizer not only presents the majority of positive opinions, but also introduces the negative perspective, which aligns better with the input polarity score. This illustrates that polarity calibration integrates polarity awareness into the model, thereby mitigating polarity bias and enhancing polarity alignment.

\subsection{Visualization Analysis}

The visualization analysis of generated summaries from different models on AmaSum dataset is shown in Figure \ref{visualization_analysis}. The x-axis is input text polarity score, which is calculated as the average polarity score of sentences in input customer reviews. The y-axis is output summary polarity score, which is calculated as the average polarity score of sentences in output summary. Each blue data point represents one product, and the ideal polarity calibration lies on the green y=x line. We observe that while human consciously try to maintain the polarity level of the input text when writing summaries, the models lacking polarity calibration tend to amplify polarity bias, resulting in excessively positive or negative summaries. The calibrated summarizer achieves improved polarity calibration when compared to the base summarizer. This underscores the effectiveness of reinforcement training in refining the model's polarity alignment.

\section{Related Work}

\noindent\textbf{Opinion Summarization} has evolved for years in the natural language processing community. \citet{erkan2004lexrank} builds a graph to extract the most representative sentences as a summary. \citet{gerani2014abstractive, di2014hybrid} rely on text planners and templates. \citet{isonuma2019unsupervised} introduces an unsupervised approach for single review summarization. \citet{brazinskas-etal-2020-unsupervised} designs an abstractive method by modeling the hierarchical continuous latent representations. \citet{iso-etal-2021-convex-aggregation} proposes an optimized combination method to encodes reviews and aggregate review embeddings. \citet{angelidis-etal-2021-extractive} maps sentences to a discrete encoding space through vector quantization and extracts the representative sentences from clusters. \citet{basu-roy-chowdhury-etal-2022-unsupervised} develops an extractive method by encoding sentences as mixtures of learned embeddings. \citet{hosking-etal-2023-attributable} proposes both extractive and abstractive models by leveraging hierarchical discrete latent space. In contrast to previous work, we aim to address the issue of amplifying polarity bias in opinion summarization models, by incorporating polarity calibration through reinforcement learning.

\vspace{2pt}

\noindent\textbf{Bias Mitigation} has garnered increasing attention in recent years \cite{lei-huang-2023-identifying, lei-huang-2023-discourse}. The majority of research to address bias mitigation focus on gender bias \cite{sun2019mitigating} or political bias \cite{lei-etal-2022-sentence}. \citet{manzini2019black} aims to detect and remove multi-class bias in word embeddings. \citet{bordia2019identifying} identifies and reduces gender bias in word-level language models. Recent work devise methods to correct linguistics bias. \citet{pryzant2020automatically, madanagopal2023reinforced} reduce linguistic bias by editing text segments such as words or sentences. \citet{liu2021transformer} introduces a transformer-based model to reduce bias by rewriting biased text. In contrast to these studies, our research investigates the issue of polarity bias in opinion summarization. Our approach aims to mitigate polarity bias and enhance polarity alignment in subjective summarization.

\vspace{2pt}

\noindent\textbf{Reinforcement Learning} has been frequently used for sequence generation tasks to mitigate exposure bias or to directly optimize task-specific evaluation metrics \cite{ranzato2015sequence, henss2015reinforcement, bahdanau2016actor, paulus2017deep, fedus2018maskgan}. In addition, reinforcement learning has been explored for a variety of natural language processing tasks such as question answering \cite{xiong2017dcn+}, knowledge graph reasoning \cite{lin2018multi}, relation extraction \cite{qin2018robust}, language generation \cite{li2016deep}, and text summarization \cite{chen2018fast}. Our work develop a reinforcement learning approach to calibrate polarity, by designing rewards for polarity bias, content preservation, and language naturality.

\section{Conclusion}

This paper focuses on opinion summarization task. We conduct an analysis of previous summarization models, which reveals their tendency to amplify the polarity bias in input text. To mitigate polarity bias and improve polarity alignment between output summary and input text, we introduce the concept of polarity calibration. A reinforcement learning approach is developed for polarity calibration, by designing three rewards for polarity distance, content semantics, and language fluency. Experiments demonstrate the effectiveness of our approach in calibrating polarity while preserving content semantics and language quality.

\section*{Limitations}

In this paper, we have presented a reinforcement learning-based approach for polarity calibration. To enhance the robustness of our method, future research should investigate the influence of various reward model configurations and alternative reward model designs on polarity calibration. Besides, the experiments focus on summarizing two specific types of opinions, product reviews and political opinions articles. To broaden the scope of our approach and assess its applicability across diverse domains, it would be valuable to examine the effectiveness of polarity calibration in other types of opinion summarization tasks.

\section*{Ethical Considerations}

This paper investigates the issue of amplifying polarity bias within subjective summarization. The polarity bias is a type of unwanted bias, which hinders the fair representation of both majority and minority opinions in summarization models. The goal of this paper is to mitigate the unwanted polarity bias and enhance polarity alignment in the opinion summarization model. The release of code and model should be leveraged to address and reduce unwanted bias, serving a broader social good.

\section*{Acknowledgements}

We would like to thank the anonymous reviewers for their valuable feedback and input.

\bibliography{anthology,custom}

\appendix

\newpage

\section{Implementation}

The polarity calibration is implemented within two steps: firstly training a base summarizer with supervised learning to equip the model with opinion summarization ability, and secondly training a calibrated summarizer through reinforcement learning to refine the model's polarity alignment.

In the supervised learning stage, the number of training epochs is set to 10. We use the AdamW \cite{loshchilov2019decoupled} as the optimizer. The weight decay is set to 1e-2. The batch size is 32. The portion of warm up phase is 0.05. The learning rate is initialized as 1e-5 and adaptively adjusted by a linear scheduler.

In the reinforcement learning stage, the weights $\alpha, \beta, \gamma$ assigned to the polarity distance reward, content preservation reward, and language naturality reward in equation (4) are 1.0, 0.5, 0.2 respectively. The weight decay is set to 1e-2. The batch size is 32. The learning rate is set to 1e-6.

\section{Evaluation of Reward Models}

The polarity score prediction model for product reviews summarization is a sentiment analysis model. A binary classifier is built to categorize the text into positive or negative class. The classifier is trained on the Amazon product reviews dataset \cite{marc_reviews}. The Precision is 0.9052, Recall is 0.9022, and F1 score is 0.9035.

The polarity score prediction model for political articles summarization is a political stance prediction model. A binary classifier is built to categorize each article into liberal or conservative stance. The model is trained on the political stance dataset AllSides \cite{baly-etal-2020-detect}. The Precision is 0.8829, Recall is 0.8906, and F1 score is 0.8864.

The content preservation reward model is a content similarity model that predicts a similarity score between two text. The STS-B semantic similarity dataset \cite{wang2018glue} annotates the similarity score for each text pair from 1 to 5. The model is trained with the mean squared error loss function to predict these scores. The Pearson correlation evaluated on the eval set is 0.9109.

The language naturality reward model is a language fluency prediction model. A binary classifier is built to predict the text into grammatical correctness or not. The model is trained on the Corpus of Linguistic Acceptability (CoLA) dataset \cite{wang2018glue}. The accuracy on the eval set is 0.8504.

\section{Human Evaluation}

\label{human_eval_questions}

The human evaluation aims to assess the generated summaries from three perspectives: polarity bias, content semantics, and language quality. Specifically, we provide the human annotators with input text and model generated summaries, and ask them the following five questions related to polarity bias, content non-hallucination, content non-redundancy, language fluency, and language coherency.

\begin{enumerate}
    \item Is the \textbf{polarity} of the generated text too positive or too negative compared to the input text? Choose 2, 1, 0, -1, -2. Scores explanation: 2 - far more positive than the target, 1 - a little more positive, 0 - very close, -1 - a little more negative, -2 - far more negative
    \item Does the \textbf{content} of the generated text hallucinate compared to the reviews? Choose 1 or 0. Scores explanation: 1 - not hallucinate, 0 - has hallucinations
    \item Is the \textbf{content} of the generated text redundant? Choose 1 or 0. Scores explanation: 1 - concise and not redundant, 0 - has redundancy content
    \item Is the \textbf{language} of the generated text fluent and grammatically correct? Choose 1 or 0. Scores explanation: 1 - fluent and grammatically correct, 0 - not fluent and has grammar errors
    \item Is the \textbf{language} of the generated text coherent? Choose 1 or 0. Scores explanation: 1 - coherent, 0 - not coherent
\end{enumerate}

\section{Prompt for ChatGPT and GPT-4}

\label{gpt-prompt}

The prompt provided into gpt-3.5-turbo and gpt-4 baselines for product reviews summarization is "Please summarize the following customer reviews text. Text: <reviews>. Summary:"

The prompt provided into gpt-3.5-turbo and gpt-4 baselines for political articles summarization is "Please summarize the following text. Text: <articles>. Summary:"

\begin{figure*}[ht]
  \centering
  \includegraphics[width = 6.3in]{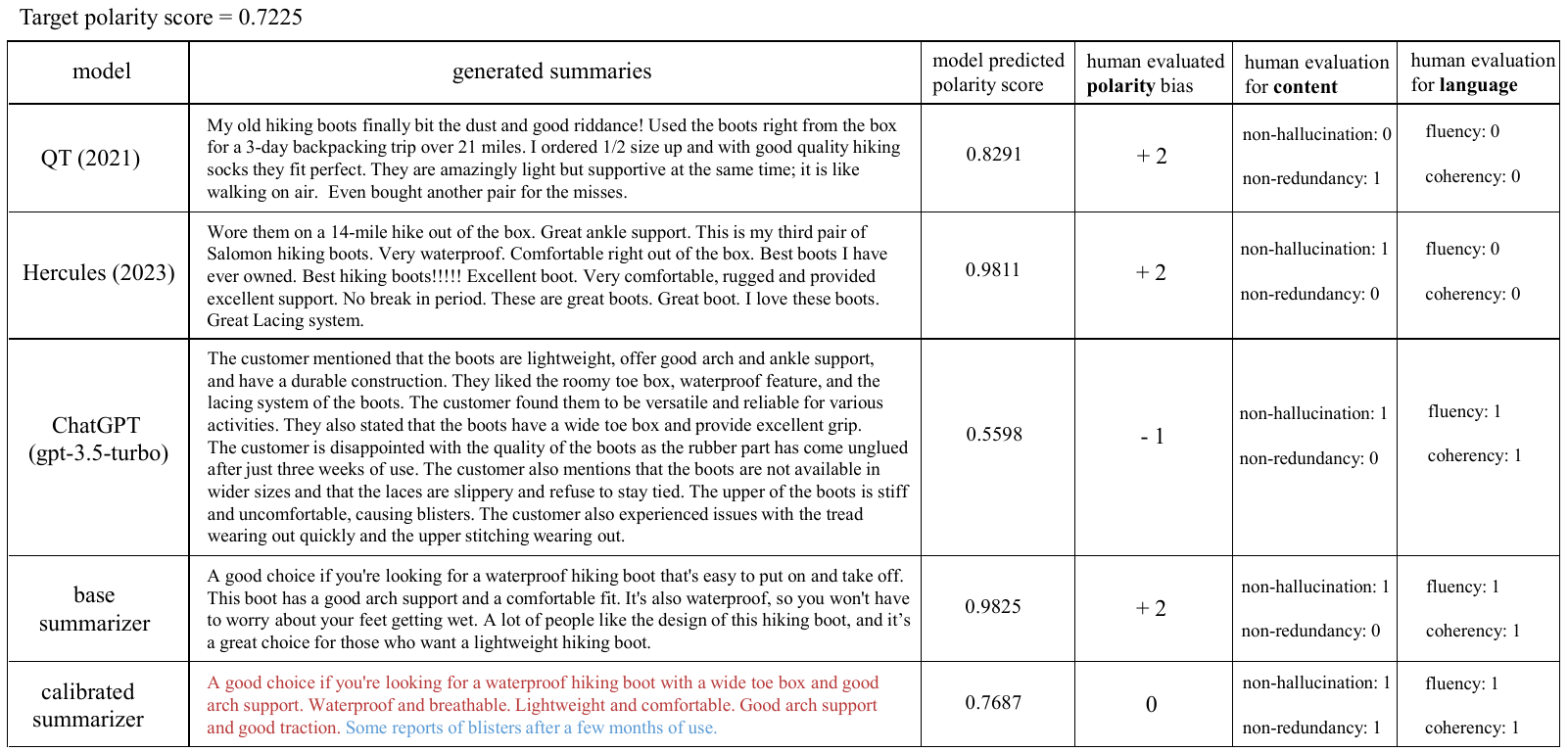}
  \caption{The qualitative analysis of generated summaries from various models.}
  \label{qualitative_analysis_full}
\end{figure*}

\section{Qualitative Analysis}

Figure \ref{qualitative_analysis_full} provides an example of generated summaries from different models. The polarity score of input reviews which is also the target polarity score of output summary is 0.7225. The polarity score of summaries generated by each model is provided. The summary generated by the calibrated summarizer has the closest polarity score with the input text. The human evaluation results for each model are also provided.

\end{document}